\documentclass[a4paper]{article}
\usepackage[latin9]{inputenc}
\usepackage{array}
\usepackage{amsmath}
\usepackage{amssymb}
\usepackage{graphicx}
\usepackage[unicode=true]
 {hyperref}
\usepackage{breakurl}

\makeatletter

\providecommand{\tabularnewline}{\\}

\usepackage{INTERSPEECH2016}

\usepackage{bm}

\ninept

\title{Contrastive Entropy: A new evaluation metric for unnormalized language models}

\def\name#1{\gdef\@name{#1\\}}
 \name{{\em Kushal Arora, Anand Rangarajan}}

\address{Dept. of Computer and Information Science and Engineering, \\
University of Florida, Gainesville, Fl, USA \\
  {\small \tt \{karora, anand\}@cise.ufl.edu}
}

\makeatother

\begin{document}
\maketitle 
\begin{abstract}
Perplexity (per word) is the most widely used metric for evaluating
language models. Despite this, there has been no dearth of criticism
for this metric. Most of these criticisms center around lack of correlation
with extrinsic metrics like word error rate (WER), dependence upon
shared vocabulary for model comparison and unsuitability for unnormalized
language model evaluation. In this paper, we address the last problem
and propose a new discriminative entropy based intrinsic metric that
works for both traditional word level models and unnormalized language
models like sentence level models. We also propose a discriminatively
trained sentence level interpretation of recurrent neural network
based language model (RNN) as an example of unnormalized sentence
level model. We demonstrate that for word level models, contrastive
entropy shows a strong correlation with perplexity. We also observe
that when trained at lower distortion levels, sentence level RNN considerably
outperforms traditional RNNs on this new metric.
\end{abstract}

\section{Introduction}

There are two standard evaluation metrics for language models: perplexity
and word error rate (WER). The simpler of these two, WER, is the percentage
of erroneously recognized words $E$ (deletions, insertions, substitutions)
to the total number of words $N$ in a speech recognition task i.e.

\begin{equation}
\mathrm{WER}=\frac{E}{N}\times100\%.
\end{equation}
The second metric, perplexity (per word), is an information theoretic
measure that evaluates the similarity between the proposed probability
distribution $m$ and the original distribution $p$. It can be computed
as an inverse of the (geometric) average probability of test set $T$
\begin{equation}
\mathrm{PPL}(T)=\frac{1}{\sqrt[n]{m(T)}}\label{eq:ppl_def}
\end{equation}
where $n$ is the number of words in the test set $T$.

In many ways, WER is a better metric. Any improvements on language
modeling benchmarks is meaningful only if they translate to improvements
in Automatic Speech Recognition (ASR) or Machine Translation. The
problem with WER is that it needs a complete ASR pipeline to evaluate.
Also, almost all benchmarking datasets are behind a pay-wall, hence
not readily available for evaluation.

Perplexity, on the other hand, is a theoretically elegant and easy
to compute metric which correlates well with WER for simpler n-gram
models. This makes PPL a good substitute for WER when evaluating n-grams
models, but for more complex language models the correlation is not
so strong \cite{iyer1997analyzing}. In addition to this, due to its
reliance on exact probabilities, perplexity is an unsuitable metric
to evaluate unnormalized models for which the partition function is
intractable. Also, when comparing two models using perplexity, they
must share the same vocabulary.

Most of the previous work done to improve upon perplexity has been
focused on achieving better correlation with WER. Iyer \emph{et al.}
\cite{iyer1997analyzing} proposed a decision tree based metric that
uses additional features like word length, POS tags and phonetic length
of words to improve the WER correlation. Chen \emph{et al. }\cite{chen1998evaluation}
proposed a new metric \emph{M-ref} which attempts to learn the likelihood
curve between WER and perplexity. Clarkson \emph{et al.} \cite{clarkson1999towards}
attempted to use entropy in conjunction with perplexity---empirically
learning the mixing coefficients. 

In this paper we focus on a different problem, the problem of extending
perplexity for unnormalized language models evaluation. We do so by
introducing a discriminative approach to language model evaluation.
Our approach is inspired by Contrastive Estimation \cite{smith2005contrastive}
and stems from the philosophical starting point that a superior language
model should be able to distinguish better between the sentence from
the test set and its deformed version. While we use an unnormalized
sentence level model as an example in this paper this technique should
work for all models where partition function is intractable, for example
unnormalized Model M and feed forward neural network language model
(NNLM) from \cite{sethy2015unnormalized} or sentence level models
like \cite{rosenfeld2001whole}, \cite{okanohara2007discriminative}
and \cite{collobert2008unified}. 

In the next section, we give a sketch derivation of perplexity that
highlights its word level model assumption. As we will be using a
sentence level language model for evaluation, we then move the probability
space to sentences and derive an expression for cross entropy rate
for sentence level models. In Section~\ref{sec:Contrastive-Perplexity},
we introduce our new discriminative metric, Contrastive Entropy, which
removes the normalization requirement associated with perplexity.
In Section~\ref{sec:Sentence-level-RNNLM}, we formulate recurrent
neural networks as sentence level language models that we use for
validation and in Section~\ref{sec:Experiments} we analyze this
new metric across various models on the Pen-TreeBank section of the
WSJ dataset. We conclude this paper by hypothesizing a better correlation
between WER and contrastive entropy based on the fact they share the
same goal of minimizing errors in prediction.

\section{Sentence level cross entropy rate\label{sec:Sentence-level-perplexity}}

The Perplexity defined in equation~(\ref{eq:ppl_def}) can also seen
as exponentiated cross entropy rate, $H(p,m)$, with cross entropy
approximated as 
\begin{equation}
H(p,m;T)=-\frac{1}{n}\log(m(T)).\label{eq:cross_ent_def}
\end{equation}
This approximation can be derived viewing language as one continuous,
infinite stream of words leading to the following expression for cross
entropy rate:

\begin{equation}
H(p,m)=\lim_{l\rightarrow\infty}-\frac{1}{l}\sum_{w_{1}^{l}\in W_{1}^{l}}p(w_{1}^{l})\log(m(w_{1}^{l})).\label{eq:cross_ent_rate_def}
\end{equation}

where $W_{1}^{l}$ is a set of all the sentences of length $l$

Now, assuming the language to be ergodic and stationary, the Shannon-McMillan-Breiman
Theorem \cite{algoet1988sandwich} states that (\ref{eq:cross_ent_rate_def})
can be approximated as a single sequence that is long enough, hence

\begin{equation}
H(p,m)=-\frac{1}{n}\log(m(w_{1}^{n})).\label{eq:ent_rate_2}
\end{equation}

Here, $w_{1}^{n}$ is the test set $T$ and $n$ being the number
of words in this test set.

In this derivation language was seen as an infinite stream of words.
If instead, we build a sample space on sentences, then we can define
the cross entropy of language as an infinite stream of sentences as
\[
H(p,m)=\lim_{l\rightarrow\infty}-\frac{1}{l}\sum_{D\in D_{1}^{l}}p(D)\log(m(D)).
\]
 $D_{1}^{l}$ here is a set of all documents containing $l$ sentences.

Now, applying the Shannon-McMillan-Breiman Theorem as we did in (\ref{eq:ent_rate_2})
and assuming that the sentences are independent and identically distributed,
we can approximate the cross entropy rate of the sentence level model
as 
\begin{flalign}
H(p,m;T)= & -\frac{1}{N}\log(m(T))\nonumber \\
 & -\frac{1}{N}\sum_{W_{n}\in T}\log(m(W_{n})).\label{eq:sent_cross_ent_rate}
\end{flalign}
where $N$ is the number of sentences in the test set $T$.

As the cross entropy still depends upon the exact probability, equation~(\ref{eq:sent_cross_ent_rate})
is still intractable. In the next section, we overcome this problem
by defining a discriminative evaluation metric which, instead of trying
to minimize the distance between the original distribution $p$ and
the proposed distribution $m$, tries to maximize the discriminative
ability of the model towards the test set from its distorted version.

\section{Contrastive Entropy and Contrastive Entropy Ratio\label{sec:Contrastive-Perplexity}}

Let $T$ be the test set. We pass this test set through a noisy channel
and let the distorted version of this test set be $\hat{T}$. We now
define the contrastive entropy rate as
\begin{eqnarray}
H_{C}(T;d) & = & H(\hat{T};d)-H(T)\nonumber \\
 & = & -\frac{1}{N}\log\left(\frac{p(\hat{T};d)}{p(T)}\right)\nonumber \\
 & = & -\frac{1}{N}\log\left(\frac{\tilde{p}(\hat{T};d)}{\tilde{p}(T)}\right).\label{eq:contrastive_entropy}
\end{eqnarray}
Here, $d$ is a measure of the distortion introduced in the test set,
$\tilde{p}$ is the unnormalized probability and $N$ is the size
of the test set, which is the cardinality of words and sentences for
word and sentence level models respectively.

The intuition behind our evaluation technique is that the distorted
test set $\hat{T}$ can be seen as an out of domain text, and that
a superior language model should be able to better discriminate in-domain
text from the language from the malformed set that are less likely
to be generated by the same language source. 

The metric proposed above still has a major drawback. It is not scale
invariant. Let's say a model \emph{M} generates a probability distribution
$m$ for test set $T$. We can simply cheat on this metric by proposing
a model that exponentiates the probability by a factor of $k$, i.e.
multiplies the entropy by factor of $k$. This limits the usefulness
of the contrastive entropy to intra-model comparison for hyper-parameter
optimization. 

We overcome this issue by reporting an additional value for each model
which we term the contrastive entropy ratio. The idea here is to choose
a distortion level as baseline, let's say 10\% and report the gain
for a higher distortion levels, for example 30\% over this baseline
distortion : 
\begin{equation}
H_{CR}(T;d_{b},d)=\frac{H_{c}(T,d)}{H_{c}(T,d_{b})}.
\end{equation}

\begin{table}[t]
\begin{centering}
\begin{tabular*}{1\columnwidth}{@{\extracolsep{\fill}}>{\raggedright}m{0.3\columnwidth}>{\raggedright}m{0.3\columnwidth}>{\raggedright}m{0.3\columnwidth}}
\hline 
$H_{C}$/$H_{CR}$ & Higher or similar $H_{CR}$ & Lower $H_{CR}$\tabularnewline
\hline 
Higher $H_{C}$ & Superior  & Scaling issues\tabularnewline
Lower $H_{C}$ & Indeterminate & Inferior\tabularnewline
\hline 
\end{tabular*}
\par\end{centering}

\centering{}\caption{Contrastive Entropy ($H_{C}$) and Contrastive Entropy Ratio ($H_{CR}$)
matrix\label{tab:H_C_vs_H_CR}}
\end{table}

Neither of the two numbers can provide a complete picture in isolation.
Contrastive entropy can be cheated upon by scaling entropy, on the
other hand, there is no guarantee that the contrastive entropy ratio
would rise faster for a better discriminative model, but together,
they balance each other out. Table~\ref{tab:H_C_vs_H_CR} shows how
to interpret these values. A model with higher contrastive entropy
and a higher or similar contrastive entropy ratio would mean that
it performs better at discriminating the good examples from the bad
ones, whereas, a larger contrastive entropy with lower ratio would
mean that models use different scales, and a higher ratio with lower
cross entropy would not mean much while comparing the two models.

\section{Sentence-level RNNLM\label{sec:Sentence-level-RNNLM}}

As the metric we proposed here benchmarks the unnormalized level models,
in this section we propose a simple sentence level language model
that we can use to show the efficacy of our metric. This new model
is simply an unfolded Recurrent Neural Network Language Model \cite{mikolov2010recurrent}
build at sentence level and trained to maximize the margin between
a valid sentence and its distorted version.

The Recurrent Neural Network based Language model can be defined recursively
using the following equations
\begin{eqnarray}
x(t) & = & \big[w(t-1){}^{T}s(t-1)^{T}\big]^{T},\label{eq:rnn_phrase}\\
s(t) & = & f(Ux(t)),\,\mathrm{and}\label{eq:rnn_nlinear}\\
y(t) & = & g(Ws(t)).\label{eq:p_w_t_w_1_t_1}
\end{eqnarray}

Equations~(\ref{eq:rnn_phrase}) and (\ref{eq:rnn_nlinear}) can
be seen as building latent space representations of phrases using
words and history and (\ref{eq:p_w_t_w_1_t_1}) can be seen as predicting
the probability of this word given the context. This phrasal representation
built in (\ref{eq:rnn_phrase}) and (\ref{eq:rnn_nlinear}) then would
be treated as the history for the next step. A standard sigmoidal
nonlinearity is used for $f$ and the probability distribution function
$g$ is a standard \emph{softmax}.

If we limit the context to sentence levels and move the probability
space to the \emph{sequence of the words} or \emph{n-grams}, equation~(\ref{eq:rnn_phrase})
and (\ref{eq:rnn_nlinear}) can be seen as composition function building
phrase $x(t)$, of the length $n$, from sub-phrase $x(t-1)$, of
the length $n-1$, and the $n$th word $w(t)$. Equation~(\ref{eq:p_w_t_w_1_t_1})
can be seen as building the unnormalized probability $\tilde{p}$
over the phrase $x(t)$. We can rephrase the equations~(\ref{eq:rnn_phrase}),
(\ref{eq:rnn_nlinear}) and (\ref{eq:p_w_t_w_1_t_1}) as 
\begin{eqnarray}
x(t) & = & f\left(U\left[\begin{array}{c}
x(t-1)\\
w(t)
\end{array}\right]\right),\,\mathrm{and}\\
y(t) & = & g(Wx(t)).
\end{eqnarray}
Here we use the standard sigmoidal non linearity for the function
$f$ and the identity function for $g$.

We now define the score of a length $N$ sentence $W$ as 
\begin{equation}
S(W)=\sum_{t=1}^{N}y(t).\label{eq:s_w}
\end{equation}

The probability of the sentence can now be modeled as an exponential
distribution 
\begin{equation}
p(W)=\frac{1}{Z}e^{-S(W)}.
\end{equation}
where $Z$ is the partition function and the contrastive entropy from
(\ref{eq:contrastive_entropy}) can be calculated as 
\begin{equation}
H_{c}(T)=1/N\sum_{W\in T}\left(S(\hat{W}_{d})-S(W)\right).
\end{equation}

where $\hat{W}_{d}$ is the distorted version of $W$ with distortion
percentage $d$.

Training is done using a contrastive criterion where we try to maximize
the distance between the in-domain sentence and its distorted version.
This formulation is similar to one followed by Collobert \emph{et
al.} \cite{collobert2008unified} and Okanohara \emph{et al. }\cite{okanohara2007discriminative}
for language modeling and by Smith and Eisner \cite{smith2005contrastive}
for POS tagging. Mathematically, we can define this pseudo discriminative
training objective as 
\begin{equation}
\theta^{\star}=\arg\min_{\theta}\sum_{d\in D}\max\left\{ 0,1-S(W_{d})+S(\hat{W}_{d})\right\} .
\end{equation}
where $\hat{W}_{d}$ is the distorted version of sentence $W_{d}$
and $\theta=(U,X,W)$ is the parameter of the model. 

This simplistic sentence level recurrent neural network model is implemented
in python using Theano \cite{bergstra+al:2010-scipy} and is available
at \emph{\href{https://github.com/kushalarora/sentenceRNN}{https://github.com/kushalarora/sentenceRNN}}.

\section{Experiments \label{sec:Experiments}}

We use the Pen Treebank dataset with the following splits and preprocessing:
Sections 0-20 were used as training data, sections 21-22 for validation
and 23-24 for testing. The training, validation and testing token
sizes are 930k, 74k and 82k respectively. The vocabulary is limited
to 10k words with all words outside this set mapped to a special token\textbf{
$<unk>$.}

\begin{figure}[!t]
\begin{centering}
\includegraphics[width=0.9\columnwidth]{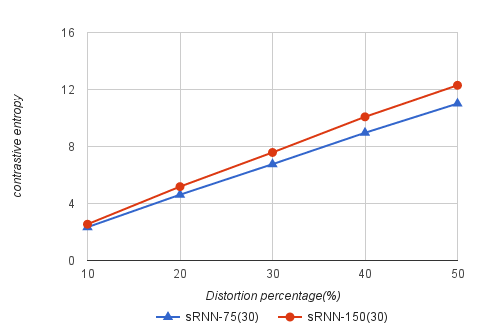}
\par\end{centering}

\caption{Test contrastive entropy monotonically increasing for with test set
distortion level. \label{fig:cppl_vs_distortion}}
\end{figure}

\begin{figure}[!h]
\begin{centering}
\includegraphics[width=0.9\columnwidth]{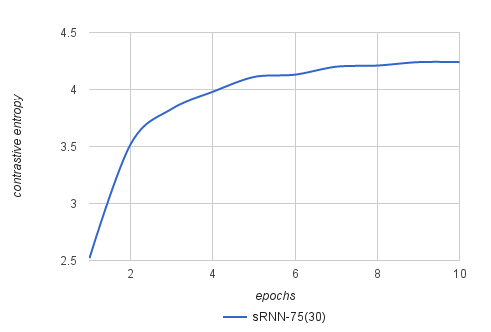}
\par\end{centering}

\caption{Training contrastive entropy monotonically increasing with epochs.
\label{fig:cppl_vs_epoch}}
\end{figure}

\begin{figure}[!h]
\centering{}\includegraphics[width=0.9\columnwidth]{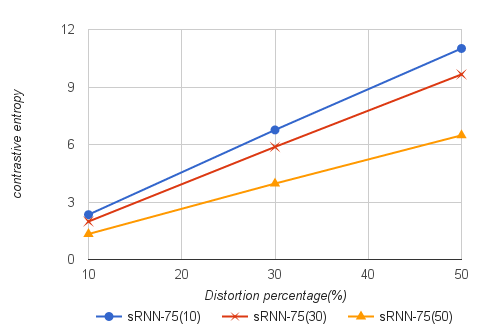}\caption{Test contrastive entropy increasing with training distortion margin.
Yellow , red and blue line represent training with distortion margin
of 50\%, 30\% and 10\% respectively. \label{fig:cppl_ppl_coer_models}}
\end{figure}

We start by examining the distortion generation mechanism. As the
evaluation includes the word level models, we need to preserve the
word count. To do this, we restrict distortions to only two types:
substitution and transpositions. For substitutions, we randomly swap
the current word in the sentence with a random word from the vocabulary.
For transposition, we randomly select a word from the same sentence
and swap it with the current one. For each word in a sentence, there
are three possible outcomes: no distortion with probability $x_{\mathcal{N}}$,
substitution with probability $x_{S}$ and transposition with probability
$x_{T}$ with $x_{\mathcal{N}}+x_{S}+x_{T}=1$.

Now, let's start by considering the sentence level RNN model proposed
in section~\ref{sec:Sentence-level-RNNLM}. For contrastive entropy
to be a good measure for sentence level models, the following assertions
should be true: i) contrastive entropy should monotonically increase
with distortions, ii) contrastive entropy of training set should go
down with each epoch, and iii) contrastive entropy should increase
with increase in training distortion margin. Figures~\ref{fig:cppl_vs_distortion},
\ref{fig:cppl_vs_epoch} and \ref{fig:cppl_ppl_coer_models} show
that the assertions made above empirically hold. We see a monotonic
increase in contrastive entropy with distortion and training distortion
margin in Figures~\ref{fig:cppl_vs_distortion} and \ref{fig:cppl_ppl_coer_models}
respectively. Figure~\ref{fig:cppl_vs_epoch} shows the contrastive
entropy increase for training data with epochs. All sentence level
RNN model referred above and elsewhere in this paper were trained
using gradient descent with learning rate of 0.1 and $\ell_{2}$ regularization
coefficient of $10^{-3}$. 

Finally, we would like to compare the standard word level baseline
models and our sentence level language model on this new metric. The
objective here is to verify the hypothesis that between two language
models, the superior one should be able to better distinguish the
test sentence from their distorted versions. This is akin to saying
that a better language model should have higher contrastive entropy
value with similar or higher cross entropy ratio. Tables~\ref{tab:H_C}
and \ref{tab:H_CR} shows the results for our experiments. The results
were generated using the open source language modeling SRILM toolkit
\cite{stolcke2002srilm} for n-gram models and the RNNLM toolkit \cite{mikolov2011rnnlm}
for the RNN language model. The RNN model used had 200 hidden layers,
with class size of 50. The sRNN-75(10) row in Tables~\ref{tab:H_C}
and \ref{tab:H_CR} indicates that the sentence level RNN model was
trained with latent space size of 75 and with training distortion
level of 10\%. All the results here were averaged over 10 runs.

\begin{table}[t]
\begin{centering}
\begin{tabular*}{1\columnwidth}{@{\extracolsep{\fill}}|>{\centering}p{0.3\columnwidth}|>{\centering}p{0.12\columnwidth}>{\centering}p{0.1\columnwidth}>{\centering}p{0.1\columnwidth}>{\centering}p{0.1\columnwidth}|}
\hline 
Model & $PPL$ & 10\% $H_{C}$ & 30\% $H_{C}$ & 50\% $H_{C}$\tabularnewline
\hline 
3-gram KN & 148.28 & 1.993 & 4.179 & 5.279\tabularnewline
5-gram KN & 141.46 & 2.021 & 4.198 & 5.308\tabularnewline
RNN & 141.31 & 2.546 & 5.339 & 6.609\tabularnewline
sRNN-75(50) & - & 1.978 & 3.961 & 6.477\tabularnewline
sRNN-75(10) & - & 2.339 & 6.759 & 11.01\tabularnewline
sRNN-150(10) & - & 2.547 & 7.581 & 12.925\tabularnewline
\hline 
\end{tabular*}
\par\end{centering}

\centering{}\caption{Contrastive entropy ($H_{C}$) and perplexity ($PPL$) for n-gram,
RNNLM and sRNN models at 10\%, 30\% and 50\% distortion levels.\label{tab:H_C}}
\end{table}

As hypothesized, the contrastive entropy rises in Table~\ref{tab:H_C}'s
columns 2 to 4 and correlates negatively with perplexity for word
level models---i.e. the models expected to do better on perplexity
do better on Contrastive entropy as well. Rows 4 to 6 compare sentence
level RNN models. Here too, as expected, sRNN trained with distortion
level of 10\% outperforms sRNN trained with distortion margin of 50\%.
Now, let's compare word level models to our sentence level model.
We can see that sRNN-75(50) performs worse compared to RNN for all
levels and worse than 3-gram and 5-gram models for 10\% and 30\%.
This can be attributed to the training distortion margin of 50\% which
encourages the sRNN to see anything with less than 50\% distortion
as in-domain sentences. On the other hand sRNN trained with distortion
level of 10\% performs the best as compared to all other models as
it has been tuned to label slightly un-grammatical sentences or ones
that have slightly un-natural structure as out of domain. 

Table~\ref{tab:H_CR} shows that scaling is not an issue for word
level models as ratios are more or less the same. Sentence level models
at 10\% distortion do better than all the word-level models on both
metrics which demonstrates their superior performance. sRNN-75(50)
is an interesting case. At test distortion level of 30\% it is clearly
inferior to all word level models as it was trained on a distortion
margin of 50\%. With 50\% test distortion the result is unclear as
it does worse on contrastive entropy but better on contrastive ratio.

\begin{table}[t]
\begin{centering}
\begin{tabular*}{1\columnwidth}{@{\extracolsep{\fill}}|>{\centering}p{0.3\columnwidth}|>{\centering}p{0.13\columnwidth}>{\centering}p{0.15\columnwidth}|}
\hline 
Model & 30\%/10\% $H_{CR}$ & 50\%/10\% $H_{CR}$\tabularnewline
\hline 
3-gram KN & 2.096 & 2.649\tabularnewline
5-gram KN & 2.077 & 2.626\tabularnewline
RNN & 2.097 & 2.596\tabularnewline
sRNN-75(50) & 2.002 & 3.275\tabularnewline
sRNN-75(10) & 2.890 & 5.257\tabularnewline
sRNN-150(10) & 2.976 & 5.074\tabularnewline
\hline 
\end{tabular*}
\par\end{centering}

\centering{}\caption{Contrastive entropy ratio ($H_{CR}$) for n-gram, RNN and sRNN models
at 30\% and 50\% distortion levels with baseline distortion level
of 10\%.\label{tab:H_CR}}
\end{table}

\section{Conclusion\label{sec:Conclusion}}

In this paper we proposed a new evaluation criteria which can be used
to evaluate unnormalized language models and showed, using examples,
its efficacy in comparing sentence level models among themselves and
to word level models. As both WER and contrastive entropy are discriminative
measures, we hypothesize that contrastive entropy should have a better
correlation with WER as compared to perplexity. 

We also proposed a discriminatively trained sentence level formulation
of recurrent neural networks which outperformed the current state
of the art RNN models on our new metric. We hypothesize that this
formulation of RNN does a better job at discriminative tasks like
lattice re-scoring as compared to standard RNN and other traditional
language modeling techniques. We conclude by restating that a metric
is meaningful only if it can measure improvements in real world applications.
Further experiments evaluating contrastive entropy's correlation with
the WER and BLEU metrics over a wide range of datasets are required
to unquestionably demonstrate the usefulness of this metric. Similarly,
to establish superior discriminative ability of sentence level RNNs
over standard RNNs, we must compare their performance on real word
discriminative tasks like \emph{n-best} list re-scoring.

\bibliographystyle{IEEEtran}
\bibliography{NLU}

\begin{thebibliography}{10}
\providecommand{\url}[1]{#1}
\csname url@samestyle\endcsname
\providecommand{\newblock}{\relax}
\providecommand{\bibinfo}[2]{#2}
\providecommand{\BIBentrySTDinterwordspacing}{\spaceskip=0pt\relax}
\providecommand{\BIBentryALTinterwordstretchfactor}{4}
\providecommand{\BIBentryALTinterwordspacing}{\spaceskip=\fontdimen2\font plus
\BIBentryALTinterwordstretchfactor\fontdimen3\font minus
  \fontdimen4\font\relax}
\providecommand{\BIBforeignlanguage}[2]{{%
\expandafter\ifx\csname l@#1\endcsname\relax
\typeout{** WARNING: IEEEtran.bst: No hyphenation pattern has been}%
\typeout{** loaded for the language `#1'. Using the pattern for}%
\typeout{** the default language instead.}%
\else
\language=\csname l@#1\endcsname
\fi
#2}}
\providecommand{\BIBdecl}{\relax}
\BIBdecl

\bibitem{iyer1997analyzing}
R.~Iyer, M.~Ostendorf, and M.~Meteer, ``Analyzing and predicting language model
  improvements,'' in \emph{Automatic Speech Recognition and Understanding,
  1997. Proceedings., 1997 IEEE Workshop on}.\hskip 1em plus 0.5em minus
  0.4em\relax IEEE, 1997, pp. 254--261.

\bibitem{chen1998evaluation}
S.~F. Chen, D.~Beeferman, and R.~Rosenfeld, ``Evaluation metrics for language
  models,'' 1998.

\bibitem{clarkson1999towards}
P.~Clarkson, T.~Robinson \emph{et~al.}, ``Towards improved language model
  evaluation measures.'' in \emph{EUROSPEECH}, 1999.

\bibitem{smith2005contrastive}
N.~A. Smith and J.~Eisner, ``Contrastive estimation: Training log-linear models
  on unlabeled data,'' in \emph{Proceedings of the 43rd Annual Meeting on
  Association for Computational Linguistics}.\hskip 1em plus 0.5em minus
  0.4em\relax Association for Computational Linguistics, 2005, pp. 354--362.

\bibitem{sethy2015unnormalized}
A.~Sethy, S.~Chen, E.~Arisoy, and B.~Ramabhadran, ``Unnormalized exponential
  and neural network language models,'' in \emph{Acoustics, Speech and Signal
  Processing (ICASSP), 2015 IEEE International Conference on}.\hskip 1em plus
  0.5em minus 0.4em\relax IEEE, 2015, pp. 5416--5420.

\bibitem{rosenfeld2001whole}
R.~Rosenfeld, S.~F. Chen, and X.~Zhu, ``Whole-sentence exponential language
  models: a vehicle for linguistic-statistical integration,'' \emph{Computer
  Speech \& Language}, vol.~15, no.~1, pp. 55--73, 2001.

\bibitem{okanohara2007discriminative}
D.~Okanohara and J.~Tsujii, ``A discriminative language model with
  pseudo-negative samples.'' in \emph{ANNUAL MEETING-ASSOCIATION FOR
  COMPUTATIONAL LINGUISTICS}, vol.~45, no.~1, 2007, p.~73.

\bibitem{collobert2008unified}
R.~Collobert and J.~Weston, ``A unified architecture for natural language
  processing: Deep neural networks with multitask learning,'' in
  \emph{Proceedings of the 25th international conference on Machine
  learning}.\hskip 1em plus 0.5em minus 0.4em\relax ACM, 2008, pp. 160--167.

\bibitem{algoet1988sandwich}
P.~H. Algoet and T.~M. Cover, ``A sandwich proof of the
  {Shannon-McMillan-Breiman} theorem,'' \emph{The annals of probability}, pp.
  899--909, 1988.

\bibitem{mikolov2010recurrent}
T.~Mikolov, M.~Karafi{\'a}t, L.~Burget, J.~Cernock{\`y}, and S.~Khudanpur,
  ``Recurrent neural network based language model.'' in \emph{INTERSPEECH},
  2010, pp. 1045--1048.

\bibitem{bergstra+al:2010-scipy}
J.~Bergstra, O.~Breuleux, F.~Bastien, P.~Lamblin, R.~Pascanu, G.~Desjardins,
  J.~Turian, D.~Warde-Farley, and Y.~Bengio, ``Theano: a {CPU} and {GPU} math
  expression compiler,'' in \emph{Proceedings of the Python for Scientific
  Computing Conference ({SciPy})}, Jun. 2010, oral Presentation.

\bibitem{stolcke2002srilm}
A.~Stolcke, ``Srilm-an extensible language modeling toolkit.'' in
  \emph{INTERSPEECH}, 2002.

\bibitem{mikolov2011rnnlm}
T.~Mikolov, S.~Kombrink, A.~Deoras, L.~Burget, and J.~Cernocky,
  ``{RNNLM}-recurrent neural network language modeling toolkit,'' in
  \emph{Proc. of the 2011 ASRU Workshop}, 2011, pp. 196--201.

\end{thebibliography}

\end{document}